\pdfoutput=1

\documentclass{article}

\usepackage[accepted]{icml2024}

\usepackage{booktabs}
\usepackage{enumitem}

\usepackage[T1]{fontenc}
\usepackage[final]{microtype}
\definecolor{mydarkblue}{rgb}{0,0.08,0.45}
\usepackage[colorlinks,allcolors=mydarkblue]{hyperref}
\usepackage{url}

\usepackage[capitalize,noabbrev]{cleveref}
\crefname{section}{Section}{\S\S}
\Crefname{section}{Section}{\S\S}
\crefname{table}{Table}{Tables}
\crefname{figure}{Figure}{Figures}
\crefname{algorithm}{Algorithm}{}
\crefname{equation}{eq.}{}
\crefname{appendix}{Appendix}{}
\crefformat{section}{Section #2#1#3}

\usepackage{pifont}
\definecolor{darkerGreen}{RGB}{0,170,0}
\newcommand\greencheck{\textcolor{darkerGreen}{\ding{52}}}
\newcommand\redcross{\textcolor{red}{\ding{55}}}
\newcommand\orangecircle{\textcolor{orange}{\ding{108}}}

\date{}

\begin{document}

\icmltitlerunning{Data Authenticity, Consent, \& Provenance for AI are all broken: what will it take to fix them?}

\twocolumn[
    \icmltitle{Data Authenticity, Consent, \& Provenance for AI are all broken:\\what will it take to fix them?}
    
    \icmlsetsymbol{equal}{*}
    
    \begin{icmlauthorlist}
    \icmlauthor{Shayne Longpre}{mit}
    \icmlauthor{Robert Mahari}{mit}
    \icmlauthor{Naana Obeng-Marnu}{mit,ccc}
    \icmlauthor{William Brannon}{mit,ccc}
    \icmlauthor{Tobin South}{mit}
    \icmlauthor{Katy Gero}{harvard}
    \icmlauthor{Sandy Pentland}{mit}
    \icmlauthor{Jad Kabbara}{mit,ccc}
    \end{icmlauthorlist}
    
    \icmlaffiliation{mit}{Media Lab, Massachusetts Institute of Technology, Cambridge, USA}
    \icmlaffiliation{ccc}{MIT Center for Constructive Communication, Cambridge, USA}
    \icmlaffiliation{harvard}{Harvard University, Cambridge, USA}
    
    \icmlcorrespondingauthor{Shayne Longpre}{slongpre@media.mit.edu}
    
    \icmlkeywords{Data Provenance, Artificial Intelligence}
    
    \vskip 0.3in
]

\printAffiliationsAndNotice{}

\begin{abstract}
\noindent
New capabilities in foundation models  are owed in large part to massive, widely-sourced, and under-documented training data collections.
Existing practices in data collection have led to challenges in tracing authenticity, verifying consent, preserving privacy, addressing representation and bias, respecting copyright, and  overall developing ethical and trustworthy foundation models.
In response, regulation is emphasizing the need for training data transparency to understand foundation models' limitations.
Based on a large-scale analysis of the foundation model training data landscape and existing solutions, we identify the missing infrastructure to facilitate responsible foundation model development practices.
We examine the current shortcomings of common tools for tracing data authenticity, consent, and documentation, and outline how policymakers, developers, and data creators can facilitate responsible foundation model development by adopting universal data provenance standards.
\end{abstract}

\section{The Need for Data Provenance}
\label{sec:introduction}
In the last decade, data from across the web, such as news, social media, and encyclopedias, has become a vital resource in general-purpose, generative AI consumer technologies\footnote{This work is scoped specifically to AI technologies often called foundation models that are (a) generative, meaning they can create potentially inauthentic text, images or other content, and (b) are typically general-purpose, and therefore tend to scrape vast, diverse distribution of web data, that are under-documented, curated, and checked for consenting use.} like GPT-4, Midjourney, and Whisper. These technologies, some with over 100 million weekly users \citep{malik2023openai}, have already begun to catalyze innovation \citep{brynjolfsson2018productivity} and scientific inquiry while starting to affect wide swaths of the economy and many everyday consumers~\citep{bommasani2021opportunities}.
These models are trained on diverse compilations of text, image, and audio data scraped from the web, synthetically generated, or hand-curated.
The resulting race to scrape, secure, and mass-produce massive collections of loosely structured data has come with consequences.

Current practices include widely sourcing and bundling data without tracking or vetting their original sources \citep{longpre2023data}, creator intentions \citep{chayka_aistealing_art_2023}, copyright and licensing status \citep{bandy2021addressing}, or even basic composition and properties \citep{dodge2021documenting}. 
The lack of transparency on this metadata and public infrastructure available to track it has led developers into ethical and legal challenges. 

Data that is used for training without significant due diligence has resulted in numerous recent, real-world problems. For instance, the LAION-5B dataset was among the most widely used text-to-image datasets on HuggingFace, before being removed after thousands of Child Sexual Abuse Material (CSAM) images were reported \citep{IWF2023AIAbuse, David2023AIDatasetCSAM}. The use of certain data sources has triggered intellectual property disputes, culminating in lawsuits against Stability AI, Midjourney, and OpenAI (see e.g., Tremblay v. OpenAI and The New York Time v. Microsoft Corporation).
Additionally, there is evidence that foundation models can leak personally identifiable information (PII) \citep{carlini2021extracting, nasr2023scalable}, generate non-consensual intimate imagery (NCII), create misinformation or deepfakes \citep{rawte2023survey,westerlund2019emergence}, and proliferate biases or discrimination \citep{buolamwiniGenderShades2018}.
Methods to retract, or ‘unlearn’, data from a model after training is complete are currently of limited reliability \citep{villaronga2018humans}. Machine unlearning methods often fail to fully remove the intended information, or can harm other aspects of a production model, disincentivizing their use \citep{Xu2023MachineUnlearning, Shaik2024ExploringMachineUnlearning}.
As a result, early choices around training data have long-term consequences, creating a pressing need for resources that allow developers to find and fully understand the benefits and risks of different training datasets.

These issues have motivated new data infrastructure and frameworks (which we survey in \cref{sec:existing}) to overcome challenges in sourcing training data responsibly. We isolate several tools in the existing ecosystem for foundation model data management that we identify as commonly used or cited in the AI literature. However, while these tools offer promising solutions, we find they neglect key aspects of the problems, lack interoperability with parallel standards/tools, or haven’t yet succeeded in reaching widespread adoption (\cref{sec:existing-discussion} and \cref{sec:standard}).

This position paper argues that no complete system for data provenance exists despite a multitude of solutions to different elements of the problem. {\bf A unified framework dedicated to the structured documentation of data properties is needed.} This requires action from multiple stakeholders:

\begin{itemize}[itemsep=0pt]
\item Data Creators: Adopt and advocate for standardized annotation practices; implement tagging and licensing for content.
\item AI Developers: Commit to providing structured data provenance documentation; actively contribute to and utilize dataset libraries in model training.
\item Regulators: Set and enforce minimum standards for data provenance for major developers and providers; provide necessary support and funding for the development of data libraries.
\item Researchers: Foster norms around data provenance in academic research; collaborate with other stakeholders in developing universal data provenance standards.
\end{itemize}

To respond to this urgent need, we outline the importance of data provenance for different stakeholders (\cref{sec:downstream}), recognition of the problem (\cref{sec:recognition}), and legal and regulatory considerations (\cref{sec:legal}). Moreover, we investigate the shortcomings in existing data infrastructure and highlight how the current set of solutions to trace data authenticity, consent, and provenance have limitations and trade-offs (\cref{sec:existing}). Finally, we outline recommendations to facilitate a standardized library for tracing critical data features, emphasize  the need for interoperable data documentation standards, and address challenges facing informed and responsible AI (\cref{sec:standard} and \cref{sec:takeaways}).

\section{Who Needs Data Provenance?}
\label{sec:downstream}
\subsection{Creators: Protecting Rights and Avoiding Harms}
\label{subsec:downstream-creators}
Artists and data creators emphasize data provenance in their pursuit of ``the three `C'''s of creative rights: compensation, control, and credit \cite{chayka_aistealing_art_2023}. Use of creators' work in computational contexts prior to the current AI trends, such as in academic research or for transformative works, followed social norms which developed over many years of practice \cite{fiesler2019creativity, klassen2022isn}. Current AI technologies have side-stepped such social norms, leaving creators' dissatisfied and unempowered. 
Creators' work is frequently used to train commercial AI models which can be used to generate competing works.
As plaintiffs in lawsuits against major AI companies (e.g. NY Times v. Microsoft and Sancton v. OpenAI) and as part of the ``Writer's Strike'' \citep{wga_negotiations_2023}, creators have raised concerns about the legality and ethics of using their data as well as the resulting effects on the creative economy~\citep{epstein2023art}.
A data transparency and provenance framework can help address these problems: it would afford creators valuable insight into how their work is used in AI, giving them an opportunity to provide consent for their data to be used, verify proper credit, and seek fair compensation in applicable cases where their data is used.

The unsettled legal status of AI training data \cite{CRS_Gen_AI_Copyright} has led to some initial compensation proposals \cite{qzShouldCreators2023}, over which creators have little control.
Creators are also rarely credited as having contributed training data, despite the occasional artist signature or watermark that slips through into generated content \cite{thevergeGettyImages2023}.
This lack of data transparency and legal clarity has led to several lawsuits against leading AI companies \cite{theVergeCopilotLawsuit2023, nytimesSarahSilverman, reutersAILawsuits2023, thevergeGettyLawsuit2023} and to calls from creators and publishers for strong transparency requirements around AI data \cite{complete_music_update_ai_act_2023, the_verge_news_outlets_demand_2023, bommasani2023foundation, bommasani2024foundation}.

The prospective benefits to creators, however, are not only about reducing harms. Greater data transparency can help users---including creators---know which model is best suited to their needs. The disclosure of whether and to what extent a model has been trained on a particular language, literary genre, or style of visual art, for example, can help artists in these media find the right AI tool for their work. 
Such disclosures also assist creators who see benefits in the use of AI technology in artistic practice~\cite{smee_ai_art_no_threat_2023} by highlighting likely gaps in model coverage and abilities and flagging potential misuse. 
As subject matter experts, creators could also be effective in recommending or providing new data sources.

\subsection{Developers: Informed Data Use}
\label{sec:subsec:downstream-developers}
AI developers have an acute interest in data and its provenance for model performance, model behavior, and for anticipating limitations and risks. While model performance generally improves with more data, the quality and diversity of data are also critical factors for reliable performance~\cite{kaplan2020scaling, hoffmann2022training}.
Model behavior tends to emulate the structure and composition of data at both the pretraining and finetuning stages~\cite{chung2022scaling, longpre2023pretrainers}.
For these reasons, developers curate specialized pretraining corpora for scientific writing~\citep{lewkowycz2022solving}, code \citep{li2022competition}, biomedical content \citep{singhal2023towards} and legal works \citep{henderson2022pile}.
As such, information about data sources and their properties informs AI model training.
Data documentation, provenance and analysis tooling have proven particularly important for helping practitioners understand very large datasets~\citep{bender-friedman-2018-data,gebru2021datasheets}. Existing datasets suffer from discoverability of issues related to biased or sexual content~\citep{David2023AIDatasetCSAM}), private data~\citep{nasr2023scalable}, copyright infringing data~\citep{Min2023SILOLM}, or non-commercially restricted data \citep{Heath2023ByteDanceOpenAI}.
Lastly, reproducibility and scientific progress more broadly are accelerated by data transparency and structured documentation~\citep{kapoor2022leakage}.

Tools to trace data origins~\citep{longpre2023data}, and provide large corpus analysis tools \cite{gao2020pile, dodge2021documenting, elazar2023whats} are increasingly relevant for more informed and responsible model development.
The Hugging Face platform has structured documentation for models and datasets~\citep{wolf2019huggingface, lhoest2021datasets}\footnote{\scriptsize{\url{https://huggingface.co/datasets/librarian-bots/dataset_cards_with_metadata}}}, while the Data Provenance Initiative~\citep{longpre2023data} has closely traced data properties, permissions, and lineage.
Model developers recognize that detailed systematic coverage of training data would enable more informed modeling, deeper analysis, and accessibility, and increase the usage of useful datasets that remain underutilized for lack of documentation. 
At the same time, data transparency provides model developers with information needed to avoid unintentional leakage of synthetic data into the training set~\citep{shumailov2023curse}.

\subsection{Science \& Scholarship: Enabling Research}
\label{subsec:downstream-scholars}
Standardizing the data ecosystem promises to enhance scientific enquiry into AI and unlock new opportunities in adjacent scientific fields.
Indeed, there is already considerable research interest in the economic impacts of generative AI \cite{McElheran2023, brynjolfsson2023generative}, from productivity improvements \cite{dellacquaNavigatingJaggedTechnological2023, pengImpactAIDeveloper2023} to labor exploitation \citep{hao2023cleaning} to market concentration \citep{vipra2023market}, and beyond.
The AI industry itself is also a subject of social science research, and its inner workings are of interest to several social science disciplines \cite{lee2023talkin, ziems2023can}. Training data, as an important part of this industry's supply chain \cite{Bommasani2023EcosystemGT, cen2023aisupply}, is central and can only be properly studied if easily accessible. Models themselves are also increasingly used in and for research, from the use of generative AI to annotate data \cite{dingGPT3GoodData2023, gilardiChatGPTOutperformsCrowd2023} --- sometimes covertly used by crowdworkers \cite{veselovsky2023artificial} --- to the rise of LLM-based agents \cite{xi2023rise} in applications as complex as simulations of entire societies \cite{horton2023large}. There are also studies of phenomena like public opinion \cite{chu2023language} based on training diets.

In all of these cases, researchers have less and less understanding of how their data is generated: annotation decisions are more opaque, pretrained models are not always fully open, and model-based agents are neither as controllable as a human confederate nor as realistic as a study participant. A better understanding of what goes into these data-generating models, along dimensions like language, country of origin and tasks present in model training data, can inform us about their strengths, weaknesses, and social biases. Such an understanding begins with systematic data collection, which a data provenance standard can enable, simplify, and make accessible to a significantly larger pool of scholars.

Scholars and researchers across disciplines thus have a strong interest in ensuring model training data is annotated for provenance in a consistent, standardized way --- before biases in it can influence future research and policy.

\subsection{Broader Impacts: Reducing Risk and Fighting Bias}
\label{subsec:downstream-society}
Societal risks from AI run the gamut from privacy violations~\cite{maiberg2023404} and exposure of personally identifying information, to systemic economic impacts and job displacement \cite{AutorNewFrontiers2022, GruetzemacherDisplacement2020}, to bias and discriminatory behavior~\cite{KuritaMeasuringBias2019, kapoor2024societal}.
These risks are fundamentally tied to the data that models are trained on \citep{longpre2023pretrainers}, which, together with the context and affordances of their applications, dictate their behavior.

Perhaps the most pressing concerns, and the ones most tightly coupled to data, relate to social bias and inequitable behavior. 
There are already prominent examples of AI systems acquiring and perpetuating the biases present in their training data~\cite{Steed2021, KuritaMeasuringBias2019, David2023AIDatasetCSAM}, especially in facial recognition systems~\cite{buolamwiniGenderShades2018, RajiSavingFace2020}. 
Social pressure can only be applied to companies' data choices if they are broadly visible and documented.
Post hoc attempts to mitigate or train out biases~\cite{ZhangMitigating2018}, or to retroactively remove contentious data sources, are per se \emph{reactive}. 
A more proactive approach could be beneficial by providing deeper insights into model training data. Recent work on data humanism~\cite{LupiDataHumanism2017} and data feminism~\cite{DIgnazioDataFeminism2023} illuminates key considerations and frameworks to make data accessible through visualization and transparent data management practices, crucial steps in giving affected communities’ agency. 
A community may, for example, wish not to be represented in or (supposedly) served by an application like facial recognition-based surveillance.

\section{Growing Interest in Data Provenance}
\label{sec:recognition}
Existing norms for tracing AI data provenance have major and increasingly widely acknowledged deficits \citep{bommasani2023foundation,longpre2023data}.
Popular AI systems do not disclose even basic information about their training data (e.g., ChatGPT, Bard, Llama 2).
The pace of innovation has prompted community calls for more systematic \citep{gebru2021datasheets,bender-friedman-2018-data,mitchell2019model} and extensive \citep{Sambasivan2021EveryoneWT, rogers-2021-changing,bandy2021addressing, dodge2021documenting, longpre2023data} data documentation.
However, these calls have resulted in uneven adoption and adherence. 
Documentation issues remain particularly acute for so-called ``datasets-of-datasets'': massive collections of hundreds of datasets where the original provenance information is often neglected or lost due to the lack of standard structures. Meanwhile, practitioners have called for greater data transparency \citep{bommasani2023foundation, narayanan2023generative}, for data supply chain and ecosystem monitoring \citep{bommasani2023ai, cmareport2023ai, khan2021new}, for content authenticity verification \cite{RosentholC2PA2022}, for detailed provenance tracing on behalf of reproducible, explainable, and trustworthy AI systems~\citep{kale2023provenance,werder2022acm}, and specifically for a standardized database to document trustworthy data \citep{lohr2023big, datanutrition2021, longpre2023data}.

Regulators and lawmakers in many countries have also shown interest in these objectives. 
The U.S. and EU have taken significant steps toward data transparency and both the EU AI Act and President Biden's recent executive order on AI\footnote{\scriptsize
\url{https://www.federalregister.gov/d/2023-24283}.} include provisions related to transparency, provenance, and the need to thoroughly understand AI models' inputs (See~\cref{sec:AI_regs}). 
The EU Act, in particular, spells out specific requirements for providers of foundation models related to training data provenance.
In addition, several recent U.S. congressional bills\footnote{\scriptsize
\url{https://www.congress.gov/bill/118th-congress/senate-bill/3312}.}\footnote{\scriptsize
\url{https://www.blumenthal.senate.gov/imo/media/doc/09072023bipartisanaiframework.pdf}} propose regulatory regimes for AI that would require data transparency. 
A voluntary code of conduct put forward by Canadian authorities \cite{canadaVoluntaryCode2023} calls on model developers to publish training data descriptions while UN bodies have recommended international regulations on data rights that enshrine transparency~\cite{UNGlobalDigitalCompact2023}.

This clear interest from both the researchers and lawmakers motivates our work on unified frameworks for data provenance and transparency. 
While such standards do not address AI risks directly, they are a key prerequisite to assessing risks and helping foster more responsible AI development.

\section{The Legal Dimension of Data Provenance}
\label{sec:legal}
Before exploring the details of a Data Provenance Standard (\cref{sec:standard}), we explore relevant legal considerations. We also outline how lawmakers can pave the way for a standard that achieves important regulatory objectives.

\subsection{Provenance and copyright}

There are two general ways in which an AI model may violate copyright interests. 
First, training a model can infringe on the copyrights of those whose works are in the training data or on the copyrights of those who created the training data corpora.\footnote{The creation of these datasets raises its own copyright issues since data is generally \emph{copied} to create them~\cite{lemley2020fair}.}
Second, specific outputs of an AI model may infringe on the copyrights associated with individual works in the training data. 

AI models sometimes produce outputs that closely resemble items in the pretraining data and may thus infringe on the rights of the creators of these works (who rarely consent to their content being used). 
It is important to underscore that although the use of pretraining data may be protected by fair use,\footnote{The U.S. doctrine of fair use allows copyrighted material to be used for purposes that are (among other requirements) distant form the original purpose and in ways that have minimal impact on the market for the original work (17 U.S.C. § 107).} this does not mean that specific output will not create copyright violations.
Meanwhile, instruction tuning, finetuning and alignment datasets are frequently used in ways not permitted by their license agreements~\cite{longpre2023data}.
These datasets contain expressive elements created for the sole purpose of training machine learning models and thus their use for this purpose is unlikely to be covered by fair use~\cite{mahari2023comment}.

A robust data provenance standard could help address many important issues related to the use of copyrighted material for AI training.
For both pretraining and finetuning, a standard data provenance framework can help mitigate legal risks and aid in the enforcement of copyright interests. Copyright infringement hinges on access to protected works\footnote{\textit{Sid \& Marty Krofft Television Prods., Inc. v. McDonald’s Corp.}, 562 F.2d 1157, 1172 (9th Cir. 1977); \textit{Art Attacks Ink, LLC v. MGA Entm’t Inc.}, 581 F.3d 1138, 1143 (9th Cir. 2009)} and thus knowing what datasets were used to train a model and what works are contained in these datasets is critical to assessing copyright issues. 
As discussed, training data is often mixed and repackaged, which complicates the task of precisely identifying what data was used to train a particular model. 
A robust framework not only helps creators assert their rights when generated outputs infringe on their copyright, but also helps model developers tune their models to avoid this infringement in the first place. 
Meanwhile, for finetuning and other curated datasets, a data provenance standard can ensure that model developers have access to accurate license information, making it easier to comply with relevant restrictions.
We note that this analysis focuses on the U.S. fair use principle and that other jurisdictions have different approaches to copyright (e.g. the EU’s Text and Data Mining Exception and Singapore's Copyright Act 2021).

\subsection{AI Regulation}
\label{sec:AI_regs}
Both the EU AI Act and President Biden's recent Executive Order on the Safe, Secure, and Trustworthy Development and Use of Artificial Intelligence directly and indirectly highlight the need for transparency for AI systems. 
Both texts require clear communication of the limitations of AI systems to consumers. 
The AI Act requires the disclosure of relevant information about training, validation, and testing datasets for high-risk AI systems and a summary of copyright protected training data used in foundation models. 
The technical specifications in the Act include specific data provenance information such as how the data was obtained, labeled, and processed. Meanwhile, the Executive Order encourages regulatory agencies to emphasize transparency requirements for AI models to protect consumers.
We note that privacy regulations like the EU's GDPR are generally less relevant in the context of finetuning data which does not generally fall under the definition of personal data per Art. 4(1) of GDPR. 

\subsection{The role of lawmakers in encouraging responsible AI practices}
This paper is a call to action for dataset creators, model developers, researchers, and lawmakers. 
By understanding the nature of AI ecosystems, lawmakers can create incentives that encourage better documentation of new datasets and audits of existing data. 
While the term “transparency” is often ill-defined in AI regulation, regulators could leverage transparency obligations to encourage model developers to record information about the datasets that have been used to train models. 
In addition, policymakers could provide funding for research related to data provenance.
Today, there are perverse incentives that prevent many companies from disclosing information about their datasets as doing so may increase the probability of legal action. 
Legal authorities could consider providing safe harbor to organizations that provide necessary information about their datasets to regulators and the public.

\section{Existing Solutions}
\label{sec:existing}

\begin{table*}[htbp]
\centering
\begin{tabular}{p{3cm}p{1.5cm}p{2.5cm}p{1cm}p{1cm}p{1cm}p{1cm}p{2.3cm}}
\toprule
\textsc{Name} & \textsc{Media} & \textsc{Coverage} & \textsc{Verify} & \textsc{Search} & \textsc{Extend} & \textsc{Attr.} & \textsc{For} \\
\midrule
Content Authenticity & Limited & Growing & \greencheck & \redcross & \redcross & \greencheck & authenticity \\
Robots.txt Consent & Webpages & Growing & \greencheck & \greencheck & \redcross & \greencheck & consent \\
Consent Registration & All & Growing & \greencheck & \greencheck & \redcross & \greencheck & consent \\
Data Standards & All & Growing (uneven) & \orangecircle & \greencheck & \greencheck & \greencheck & data properties \\
Common Crawl & Webpages & Wide & \greencheck & \greencheck & \redcross & \redcross & data properties \\
Hugging Face & Limited & Growing (uneven) & \redcross & \orangecircle & \greencheck & \orangecircle & data properties \\
Data Prov. Init. & Limited & Limited & \greencheck & \greencheck & \greencheck & \greencheck & data properties \\
\bottomrule
\end{tabular}
\caption{\textbf{Comparison of data provenance interventions.} A summary across interventions of the current scope of data coverage, the ability to verify origins of data or metadata claims, the ability to search for data using metadata, whether the standard is extensible, and the ability for symbolic attribution. We see that content authenticity techniques are usually embedded into or with the data itself, but have limited scope of where they can be applied, and cannot easily be extended or searched over. 
Consent opt-in has not organized around one standard yet, and only addresses the consent problem in isolation.
Data Provenance Standards and their associated libraries can provide significantly more information, extensibility, and structure that is machine readable or searchable. The challenge arises in maintaining them, and ensuring the metadata is verifiable and accurate.}
\label{tab:standards}
\end{table*}

No complete system for data provenance exists. Instead, there is a patchwork of solutions to different elements of the problem. We identify four broad categories of these: \emph{content authenticity techniques} that attach themselves to data as it spreads, \emph{opt-in \& opt-out tools} that allow content creators to register how content should be used, \emph{data provenance standards} that allow dataset creators to document information about datasets, and \emph{data provenance libraries} that aggregate information on datasets. 
These interventions imbue unstructured data with more attribution, authenticity, and navigability in machine-interpretable formats, but none is a complete solution to the challenge of data provenance. 

\subsection{Content authenticity techniques}
\label{content-authenticity}

The growing concern for manipulated media, disinformation, and deepfakes has spurred methods that attempt to embed provenance information directly alongside or into data. 
In this way, a downstream user can ascertain the data’s source and authenticity to avoid copyright issues, ensure academic integrity, or for journalism and fact-finding.

\textbf{Coalition for Content Provenance and Authenticity (C2PA).}
A prominent example of content authenticity is the C2PA, a partnership between Adobe, Microsoft, and dozens of other corporations to design a specification that ``addresses the prevalence of misleading information online through the development of technical standards for certifying the source and history (or provenance) of media content.''\footnote{\scriptsize\url{https://c2pa.org/}}
To this end, verifiable information may be cryptographically embedded into images, videos, audio, and some types of documents in a way that is difficult to remove and that makes tampering evident.

\textbf{Digital Watermarks for AI.}
Digital watermarks have historically been used in visual and audio content. For applications of AI, these watermarks are embedded into machine-generated content~\citep{736686}, and more recently into machine-generated text \citep{kirchenbauer2023watermark}.
While these technologies hold promise, they are vulnerable to removal \citep{zhang2023watermarks}, especially for text media \citep{sadasivan2023can}.

\subsection{Opt-In \& Opt-Out}
\label{opt-in}

\textbf{Robots.txt}
The Robots Exclusion Standard uses a robots.txt file in a website’s directory to indicate to crawlers (e.g., from search engines) which parts of the website the webmaster would like to include and exclude from search indexing. 
Though this protocol lacks enforcement mechanisms, major search engines have historically respected it~\citep{KellerWarso2023MLTrainingOptOut}.
Recent proposals extend this idea to AI data, including learners.txt \citep{ippolitodonottrain}, ai.txt\footnote{\scriptsize\url{https://site.spawning.ai/spawning-ai-txt}}, or ``noai'' tag by artists \citep{DeviantArt2023AIDatasetsOptOut}.
In response, Google and OpenAI have instantiated their own versions, ``User Agents'' and ``GPTBots'', giving websites an avenue to implement opt-out standards.\footnote{\scriptsize\url{https://developers.google.com/search/docs/crawling-indexing/overview-google-crawlers},\\ \url{https://platform.openai.com/docs/gptbot}}
So far, none of these has been widely adopted.
While a robots.txt-type implementation signals a website’s preferences as an opt-in/opt-out binary, it fails to provide a more nuanced spectrum of preferences (e.g., only non-commercial open-source models may train on my data), as well as other useful metadata.

\textbf{Consent Registration.}
Organizations, such Spawning AI, are attempting to build infrastructure for the ``consent layer'' of AI data.\footnote{\scriptsize\url{https://api.spawning.ai/spawning-api}}
This involves sourcing opt-in and opt-out information from data creators, and compiling this into searchable databases (e.g., their Do Not Train Registry\footnote{\scriptsize\url{https://haveibeentrained.com/}}).
This approach obtains consent directly from creators rather than web hosts, but its granularity increases the burden associated with compiling these consent databases.

\subsection{Dataset Provenance Standards}
\label{data-prov-standards}

Beyond data authenticity and consent, standards for broader data documentation have been proposed to resolve the many other challenges including data privacy, sensitive content, licenses, source and temporal metadata, as well as relevance for training.

\textbf{Datasheets, Statements \& Cards.}
Datasheets \citep{gebru2021datasheets}, data statements \citep{bender-friedman-2018-data}, and data cards \citep{pushkarna2022data} propose documentation standards for AI datasets.
Each of these standardizes documentation of AI dataset creators, annotators, language and content composition, innate biases, collection and curation processes, uses, distribution, and maintenance. Though unevenly adopted, these efforts are widely recognized for improving scientific reproducibility and responsible AI development~\citep{boyd2021datasheets}.

\textbf{Data Nutrition Labels.}
Based on the FDA's Nutrition Facts label, a Data Nutrition Label\footnote{\scriptsize\url{https://datanutrition.org/}} automates data documentation using a registration form with 55 mostly free-text question responses.

\textbf{Data \& Trust Alliance's Data Provenance Standard.}
The Data \& Trust Alliance Data Provenance Standard\footnote{\scriptsize\url{https://dataandtrustalliance.org/our-initiatives/data-provenance-standards}} is the product of joint data documentation efforts from 19 corporations, including IBM, Nike, Mastercard, Walmart, Pfizer, and UPS. 
Motivated by the absence of an ``industry-wide method to gauge trustworthiness of data based on how it was sourced'', this standard assimilates a wide variety of industry documentation needs into a succinct structured format. 
The standard provides structured documentation analogous to Nutrition Labels, but also provides a way to trace data lineages.

\subsection{Data Provenance Libraries}
\label{data-prov-lib}

While content authenticity is embedded in the data, data documentation and provenance standards need to be aggregated in libraries for searching, filtering, and machine navigation. 
Prior work has formalized and even operationalized data governance standards~\citep{jernite2022data,pistilli2023stronger,desai2023archival,PorciunculaChapelle2022Datasphere}, but only a few efforts have gained traction in guiding AI development.

\textbf{Common Crawl.}
For pretraining text data, models routinely rely on the Common Crawl.\footnote{\scriptsize\url{https://commoncrawl.org/}}
For instance, sixty percent of GPT-3 training data is from Common Crawl~\citep{browngpt3}.
This resource provides URLs, scrape times, and the raw documents, following the robots.txt exclusion protocol, and even adheres to requests for removal (e.g., the New York Times requested their data removed from the repository \citep{BarrHays2023NYTimesAI}). 
This free library of crawled web data has wide adoption, is accurate, and has comprehensive web coverage, but it provides limited metadata.

\textbf{Hugging Face Datasets.}
Hugging Face Datasets has become a widely adopted data library for AI \citep{lhoest2021datasets}.
It has integrated data cards and Spawning’s data consent information into datasets to encourage documentation and consent filtering. 
While its coverage of AI data is vast, the documentation is uneven and often incorrect, as this information is loosely crowdsourced. Hugging Face data cards also have limited structure and searchability in the current API.

\textbf{The Data Provenance Initiative.}
The Data Provenance Initiative\footnote{\scriptsize\url{www.dataprovenance.org}} is a joint effort by AI and legal experts to contribute comprehensive and accurate structured information around the most popular textual datasets in AI, including their lineage of sources, licenses, creators, and characteristics~\citep{longpre2023data}.
This provides a richer and more accurate collection of annotations and provenance for searchable, filterable, and standardized datasets. 
However, this accuracy requires expert human labor, limiting its scale compared to Hugging Face or Common Crawl.

\subsection{Discussion \& Trade-Offs of Interventions}
\label{sec:existing-discussion}

Each category of interventions above targets different problems and comes with clear trade-offs in their benefits and limitations, as illustrated in \cref{tab:standards}.
For instance, content authenticity techniques offer built-in and verifiable provenance. 
However, they only authenticate the source or veracity of the data, without covering (or being easily extensible to) other important metadata for AI applications like copyright, privacy, bias considerations, characteristics, intended uses, or complex lineage information.
See \cref{tab:metadata} for a more comprehensive list of the different metadata that has become important to AI development in recent years.
Content authenticity techniques also apply primarily to atomic units of data, like individual images, recordings, or text files, rather than derivations or compilations, which are increasingly common for multimodal AI training.

Proposals to extend robots.txt or register data opt-in/out are designed to facilitate creator consent, but each AI company requires custom code for their own scrapers, and many AI developers may still ignore these guidelines.

On the other end of the spectrum of data richness are standards like Datasheets, Data Nutrition Labels, or The Data \& Trust Alliance's Data Provenance Standard, which encode more fine-grained information, but at the expense of accuracy and adoption incentives.
For instance, consider the three data libraries discussed in \cref{data-prov-lib}.
These three libraries trade off a spectrum of (a) the coverage of data, (b) the depth of provenance documentation, and (c) the accuracy of collected metadata.
Common Crawl is accurate with wide coverage, but not detailed. 
Hugging Face can be inaccurate with varying levels of detail, but it has extensive coverage. 
The Data Provenance Initiative is highly accurate and detailed but is currently limited in scope.

Clearly, authenticity techniques, data consent mechanisms, and data provenance standards are complementary, and each conveys distinct and important information to AI developers.
Unifying these frameworks into a standardized data infrastructure layer holds tremendous promise and is a precondition to solving the many problems of ethical, legal, and sustainable AI.

\section{Elements of a Data Provenance Standard}
\label{sec:standard}

\begin{table*}[htbp]
\centering
\begin{tabular}{p{4cm}p{12.2cm}}
\toprule
\textbf{\textsc{Metadata}} & \textbf{\textsc{Definition}} \\
\midrule
\textbf{Source Authenticity} & \small The authenticity of the data (including digital watermarks and embedded authenticity). \citep{rawte2023survey,westerlund2019emergence} \\
\midrule
\textbf{Consent \& Use Restrictions} & \small What uses or audiences the creators consent to. \citep{chayka_aistealing_art_2023,epstein2023art} \\
\midrule
\textbf{Data Types} & \small The data types (text, images, tabular, etc.) and their digital formats. \\
\addlinespace[2pt]
\textbf{Source Lineage} & \small Links to data sources from which this was assembled or derived. \citep{buolamwiniGenderShades2018} \\
\addlinespace[2pt]
\textbf{Generation Methods} & \small How the data was created, by which people, organizations, and/or models. \citep{longpre2023data} \\
\addlinespace[2pt]
\textbf{Temporal Information} & \small When the data was created, collected, released, or edited. \citep{luu2022time,longpre2023pretrainers} \\
\addlinespace[2pt]
\addlinespace[2pt]
\textbf{Private information} & \small If personally identifiable or private information is present. \citep{carlini2021extracting, nasr2023scalable} \\
\addlinespace[2pt]
\textbf{Legal considerations} & \small What legal information is attached to the data.\\
\addlinespace
\ \ \small \textbf{Intellectual property} & \small Copyright, trademark, or patent information attached. \citep{tremblay2023openai} \\
\ \ \small \textbf{Terms} & \small Associated terms of use or service accessing the data. \citep{Heath2023ByteDanceOpenAI} \\
\ \ \small \textbf{Regulations} & \small Relevant regulations, though they may be jurisdiction-dependent (e.g., the EU AI Act). \\
\addlinespace
\textbf{Characteristics} & \small An extensible set of properties for the data, relevant to different applications. \\
\addlinespace
\ \ \small \textbf{Dimensions} & \small Size of the data in measurable units. \\
\ \ \small \textbf{Sensitive content} & \small If offensive, toxic, or graphic content is present. \citep{David2023AIDatasetCSAM,IWF2023AIAbuse} \\
\ \ \small \textbf{Quality metrics} & \small Metrics associated to quality measures of the data. \citep{longpre2023pretrainers} \\
\addlinespace
\textbf{Metadata Contributors} & \small A version controlled log of edits and responsible contributors to this metadata. \\
\bottomrule
\end{tabular}
\caption{\textbf{Relevant data properties} for facilitating authenticity, consent, and informed use of AI data. Each row provides a citation motivating the significance of that data property. A unified data standard would be able to codify these information pieces.
}
\label{tab:metadata}
\end{table*}

Model developers, data creators, and the public could all benefit from transparency into AI data.
A standard data provenance framework could address these diverse needs, but existing solutions tend to address different transparency problems \emph{in isolation}. 
Instead of proposing a new standard, we highlight how existing standards can be unified to effectively address the range of relevant challenges.
\emph{A unified data provenance framework should be:}

\vspace{-1mm}
\begin{enumerate}\itemsep0em
    \item \textbf{Modality and source agnostic}: An effective standard for tracing wide-ranging data provenance should not be limited to certain modalities (e.g. text, images, video, or audio), or sources.
    \item \textbf{Verifiable}: The metadata can be verified, or its reliability assessed. Although metadata will inevitably contain errors, editing systems (like the one used by Wikipedia) or provenance confirmations created by data creators help to provide transparency and consistency.
    \item \textbf{Structured}: Structured information should be searchable, filterable, and composable, so automated tools can navigate the data, and the qualities of combined datasets can easily be inferred from merging their structured properties (e.g. combining license types).
    \item \textbf{Extensible and adaptable}: An extensible framework is adaptable to new types of metadata that emerge over time as well as to various jurisdiction-specific transparency requirements. 
    \item \textbf{Symbolically attributable}: Relevant data sources should be attributed, even as datasets are repackaged and compiled. Codifying the lineage of sources allows the resulting properties to be determined by traversing the web of data ``ancestors''.
\end{enumerate}

\section{Key Lessons for Data Provenance}
\label{sec:takeaways}
Existing data provenance solutions are piece-meal. Without a robust, well-resourced data provenance framework, developers will struggle to accurately identify and evaluate the safety, copyright implications, and relevance of datasets from a dizzying array of possibilities. 
Data creators will similarly struggle to identify how and where their content is. Without dataset provenance standards and documentation, creating such a framework will become increasingly difficult and ultimately untenable. 
While each existing solution provides important insights into the data ecosystem, a robust framework to attach metadata to datasets is needed to track how datasets are mixed, complied and used. 
We advocate for a set of actions that different stakeholders can take to make data authenticity, consent, and provenance more robust to future challenges.

\textbf{New work should seek to unify, or make interoperable, data infrastructure for authenticity and consent with other important documentation for privacy, legality, and relevance.}
Solutions to these problems are being developed in isolation, but trustworthy and responsible AI requires assessing these factors together.
Among others, \citet{pistilli2023stronger} underscore the ``necessity of joint consideration of the ethical, legal, and technical in AI ethics frameworks to be used on a larger scale to govern AI systems.''

\textbf{For Policymakers:}
Regulators play a pivotal role in shaping the future of AI through policy and guidelines. 
A data-centric approach to AI regulation can help identify and mitigate key risks. 
Policymakers can provide funding for research related to data provenance and a centralized effort to document and build provenance infrastructure. 
Currently, perverse legal incentives inhibit companies from disclosing information about their data. 
Regulators should consider legal or legislative incentives for organizations to provide necessary data transparency, as they have in the DSA for social media platforms and require standardized documentation as part of AI transparency obligations. 
These types of incentives can foster universal and interoperable standards for data authenticity, consent, and provenance.

\textbf{For AI Developers:}
AI developers are at the forefront of creating AI models and thus bear significant responsibility in ensuring ethical AI practices. It is crucial for developers to prioritize documentation responsibilities and to make public the provenance of their training data. 
When there are compelling business reasons for confidentiality, at the very least, they should publish aggregate statistics about the data provenance. This level of transparency is essential for building trust with users and the wider community and for fostering a responsible AI ecosystem.

\textbf{For Data Creators \& Compilers:}
Creators of training data play a critical role in the AI development process. It is imperative for these creators to meticulously document not only consent criteria, but the provenance of their data, including the sources and processing. Repositories and databases to register this information are already available. Such detailed documentation will significantly aid AI developers in respecting underlying rights and in understanding the nature of the data they use.

\textbf{For the Research Community:}
The research community is uniquely positioned to set norms and standards around provenance disclosure.
This could involve incorporating provenance disclosure as a requirement during research publication which would complement efforts like reproducibility checklists~\cite{rogers2021just} and ultimately help foster scientific progress.

In practice, it would be hard for individual stakeholder groups working in isolation to succeed in developing a data provenance framework. Instead, we believe that an overarching multi-stakeholder collaboration could play a key role in facilitating efforts and operations needed to develop such a standard. A consortium with representatives from each stakeholder group could coordinate the activities of each group and advocate for the adoption of a specific standard. This approach would mirror the W3 Consortium which played a central role in the development of standards for the World Wide Web, or The Data \& Trust Alliance, a not-for-profit consortium that brings together businesses and institutions across multiple industries to develop and adopt responsible data and AI practices. 

\section{Conclusion}
We underscore the urgent need for a standardized data provenance framework to address the complex challenges of AI development. The proliferation of AI models, their diverse training data sources and associated ethical, legal, and transparency concerns have culminated in a critical need for a comprehensive approach to data documentation.

Several data provenance solutions exist, such as content authenticity techniques, opt-in/opt-out mechanisms, and data documentation standards. However, each of these addresses specific aspects of a broader issue and functions in isolation. A unified data provenance framework is needed to establish an ecosystem where data authenticity, consent, privacy, legality, and relevance are holistically considered and managed.

The successful implementation of such a framework requires concerted efforts from all stakeholders in the AI field. This includes creators, who need to be empowered to tag and license their content; developers, who must adopt standards for data provenance and contribute to dataset libraries; and policymakers, who should establish transparency standards and fund the documentation and construction of data libraries.

The key takeaway is the interdependence of transparency solutions: without robust data provenance libraries, it is challenging for developers to find and evaluate datasets comprehensively. Conversely, without standardized documentation and metadata attachment to data as it travels across the web, tracking and downstream use become unfeasible. This effort requires the active participation and commitment of all involved parties to create a sustainable and trustworthy AI ecosystem.

\section*{Impact Statement}
A standard data provenance framework could have far reaching positive impacts on responsible AI development, as outlined throughout.
However, such a standard is challenging to design and adopt, as evidenced by the fact that it does not exist yet despite numerous calls from diverse stakeholders.
It is important to underscore that while a well designed standard could further important social objectives, a poorly designed standard could further entrench problematic practices: an overly onerous framework could be adopted or worse impose excessive costs on under resourced researchers and developers, further benefiting large corporate AI developers.
A standard that fails to capture important information about biases and limitations could also provide a false sense of security, and encourage the use of data that is not fit for purpose.
Finally, a data standard cannot directly address most AI harms or risks and while it can help expose them it is not a substitute for well-informed AI policies.

\section*{Acknowledgements}
We would like to thank Kevin Klyman, Yacine Jernite, Hanlin Zhang, Kristina Podnar, and Saira Jesani for their generous advice and feedback.

\clearpage
\bibliographystyle{icml2024}
\bibliography{main}

\end{document}